\documentclass[graybox, envcountchap]{svmult}

\usepackage{mathptmx}        
\usepackage{helvet}          
\usepackage{courier}         

\usepackage{makeidx}         
\usepackage{graphicx}        
\usepackage{multicol}        
\usepackage[bottom]{footmisc}

\makeindex             

\begin{document}

\frontmatter

%
%
%

\begin{dedication}
Use the template \emph{dedic.tex} together with the Springer document class SVMono for monograph-type books or SVMult for contributed volumes to style a quotation or a dedication\index{dedication} at the very beginning of your book in the Springer layout
\end{dedication}

%
%

\foreword

Use the template \textit{foreword.tex} together with the Springer document class SVMono (monograph-type books) or SVMult (edited books) to style your foreword\index{foreword} in the Springer layout. 

The foreword covers introductory remarks preceding the text of a book that are written by a \textit{person other than the author or editor} of the book. If applicable, the foreword precedes the preface which is written by the author or editor of the book.

\vspace{\baselineskip}
\begin{flushright}\noindent
Place, month year\hfill {\it Firstname  Surname}\\
\end{flushright}

%
%

\preface

Use the template \emph{preface.tex} together with the Springer document class SVMono (monograph-type books) or SVMult (edited books) to style your preface in the Springer layout.

A preface\index{preface} is a book's preliminary statement, usually written by the \textit{author or editor} of a work, which states its origin, scope, purpose, plan, and intended audience, and which sometimes includes afterthoughts and acknowledgments of assistance. 

When written by a person other than the author, it is called a foreword. The preface or foreword is distinct from the introduction, which deals with the subject of the work.

Customarily \textit{acknowledgments} are included as last part of the preface.

\vspace{\baselineskip}
\begin{flushright}\noindent
Place(s),\hfill {\it Firstname  Surname}\\
month year\hfill {\it Firstname  Surname}\\
\end{flushright}

%
%

\extrachap{Acknowledgements}

Use the template \emph{acknow.tex} together with the Springer document class SVMono (monograph-type books) or SVMult (edited books) if you prefer to set your acknowledgement section as a separate chapter instead of including it as last part of your preface.

\tableofcontents
%
%
%
\contributors

\begin{thecontriblist}
Firstname Surname
\at ABC Institute, 123 Prime Street, Daisy Town, NA 01234, USA, \email{smith@smith.edu}
\and
Firstname Surname
\at XYZ Institute, Technical University, Albert-Schweitzer-Str. 34, 1000 Berlin, Germany, \email{meier@tu.edu}
\end{thecontriblist}
%
%

\extrachap{Acronyms}

Use the template \emph{acronym.tex} together with the Springer document class SVMono (monograph-type books) or SVMult (edited books) to style your list(s) of abbreviations or symbols in the Springer layout.

Lists of abbreviations\index{acronyms, list of}, symbols\index{symbols, list of} and the like are easily formatted with the help of the Springer-enhanced \verb|description| environment.

\begin{description}[CABR]
\item[ABC]{Spelled-out abbreviation and definition}
\item[BABI]{Spelled-out abbreviation and definition}
\item[CABR]{Spelled-out abbreviation and definition}
\end{description}

\mainmatter
%
%
%

\begin{partbacktext}
\part{Part Title}
\noindent Use the template \emph{part.tex} together with the Springer document class SVMono (monograph-type books) or SVMult (edited books) to style your part title page and, if desired, a short introductory text (maximum one page) on its verso page in the Springer layout.

\end{partbacktext}

\title*{ Active Face Frontalization using Commodity Unmanned Aerial Vehicles}

\author{ 
N. Lakshminarayana, Y.  Liu, K. Dantu, V. Govindaraju, N. Napp
}

\institute{Computer Science and Engineering, University at Buffalo\\ \email{
{nagashri, yifangli, kdantu, govind, nnapp}@buffalo.edu} }
%
%
\maketitle

\vspace{-90 pt}
\abstract{
This paper describes a system by which Unmanned Aerial Vehicles (UAVs) can gather high-quality face images that can be used in biometric identification tasks. Success in face-based identification depends in large part on the image quality, and a major factor is how frontal the view is. Face recognition software pipelines can improve identification rates by synthesizing frontal views from non-frontal views by a process call {\em frontalization}.  Here we exploit the high mobility of UAVs to actively gather frontal images using components of a synthetic frontalization pipeline. We define a frontalization error and show that it can be used to guide an UAVs to capture frontal views. Further, we show that the resulting image stream improves matching quality of  a typical face recognition similarity metric. The system is implemented using an off-the-shelf hardware and software components and can be easily transfered to any ROS enabled UAVs. 
}

\section{Introduction}
\label{sec:intro}

Surveillance in remote areas is difficult due to the difficulty and cost of building and operating dense camera networks over large areas. Instead, Unmanned Aerial Vehicles (UAVs) can be used as mobile sensors to increase coverage~\cite{hsu2015face}, to identify, and to follow individuals. Biometrics, i.e. consisting biological signals, of a person have proven to be an efficient means of automated identification~\cite{Jain:2000:BI:328236.328110}. Visual biometrics such as faces, fingerprints, and iris patters, are highly discriminative. However such modalities show great appearance variations in nature. For example,the pixel level comparison of facial images from the same person can look rather different while the biometric signal  should be constant. An adaptive biometric system could reduce the efforts for capturing and encoding all such variations. UAVs being highly mobile can be controlled to obtain high quality biometric data. Typically, pilots in ground stations control UAVs remotely to acquire high quality data. However, this requires extensive human effort and places constraints on scaling up such approaches, which conflicts with the goal of covering large areas. Automation of the tracking and UAV networking could potentially change this~\cite{autodrone}. 

In the past, military has employed "Drone stares" in war zones and border areas to monitor any malicious activities~\cite{doi:10.1177/1362480610396650}. Beyond  surveillance, UAVs have also been used for search and rescue operations~\cite{searchnrescue}~\cite{geo}.

Autonomous UAVs that track humans have to detect and identify individuals efficiently. Among various biometrics, face is one of the primary modalities for human identification. However, the video quality of  aerial imagery poses severe limitations to perform facial recognition. For example~\cite{doi:10.1177/1362480610396650} lists instances of "collateral damage" caused due to the ambiguity in identifying the person of interest from the aerial imagery. 

In the biometrics community facial analysis based on facial images from print media, on-line photo postings, or other data sources which were captured incidentally are called a unconstrained faces.
Such data show huge appearance variability due to change in pose, lighting, and occlusions. Such inconsistencies degrade the quality of automatic face detection and analysis. Although humans can efficiently recognize such pose variant faces, for a computer to train and learn on such noisy representation requires massive data and a considerable amount time. 

A common approach of face recognition pipelines is to synthesize a frontal view of the face using 3D reconstruction methods from a non-frontal image. However since UAVs can be used as active sensors in biometrics to obtain a optimized view of face, we propose {\em active face frontalization}, in which real frontalized data is gather rather than synthesized from non-frontal views. 

Besides getting better recognition, frontal images of faces are more aesthetic and can be used in the UAVs photography technology popular these days.  
We define {\em active frontalisation}  based on the existing frontalisation methods.
The contributions of this paper are: Firstly, we derive a frontalization error on the pose variant face based on existing virtual fontalization techniques. We show that this metric can be efficiently used to adjust the flight path to obtain an improved view of face. Secondly, we evaluate face recognition performance of the resulting views using a similarity measure on the features obtained from a deep convolutional neural network (CNN) as commonly done in state-of-the-art biometric identification.
We present an implementation of this approach using inexpensive off-the-shelf components. 

The resulting system focuses on frontalization and is currently limited by several practical limitations related to safety and sensor quality. Autonomous UAVs creates safety hazards. Therefore, we chose a relatively safe UAV over more sophisticated but dangerous ones. Further the lack of high quality cameras and sensors in an off-the-shelf UAV needs to be overcome by using robust algorithms.

More generally, the images from UAVs are not optimal for biometric analysis.
The faces seen by the UAVs have two crucial issues: long range and lateral poses. Generally the perspective of UAVs capture faces at a distance and are not detectable. In the following section~\ref{hum det}  we describe a simple person following algorithm to direct the UAVs into the range of face detection. Section \ref{visibility}  elaborates the method used for assessment of the face quality with regard to the pose. In Section \ref{controller}, we give an overview of the controller.

\section{Related Work}
\label{sec:relwork}
\vspace{-0.4cm}
Human detection from aerial imagery was performed using thermal signatures in early days~\cite{doi}~\cite{4526559}. However due to popularity of computer vision based techniques, visual feedback from the sensors were used to track and follow humans efficiently. Pestana~\cite{pestana2014computer} investigated whether visual based object tracking and following is reliable using a cheap GPS-denied UAVs while assuring safety. This work demonstrated that current tracking algorithms, such as OpenTLD~\cite{kalal2012tracking}, are reliable to work on UAVs platform, and the proposed architecture has been able to follow unmarked targets of different sizes and from a wide range of distances even when occlusions happened. Although researchers achieved good performance for tracking and following humans, identification was still a challenging task. 
Unfortunately neither thermal images nor detectors were sufficient for identification of people. Therefore researchers started analyzing the feasibility of using face recognition on UAVs. 

Unlike static cameras or high resolution cameras, UAVs have disturbance, resolution limitation and complex flying environments (different illumination and weather, indoor and outdoor), current state-of-the-art face recognition methods may have limits and issues while they are applied on UAVs, therefore, we need to know how well face recognition perform on UAVs given different altitude, distance and angle. HSU and CHEN~\cite{hsu2015face} investigated the capability of two face recognition services ($Face^{++}$ ~\cite{facelink} and ReKognition~\cite{Orbeus}) in face detection and face recognition with different altitudes, distances, and angles of depression. The results showed that the present face recognition technologies are able to perform adequately on UAVs.

Another study was made on analyzing the face quality on various platforms~\cite{Korshunov:2011:VQF:2000486.2000488}. They based their study on three commercially available  face analysis algorithms 1)  ViolaJones
[Viola and Jones 2001] and Rowley [Rowley et al. 1998] face detection algorithms; 2) QDA-based face recognition algorithm [Lu et al. 2003]; and 3) CAMSHIFT [Bradski 1998] face tracking
algorithm. It was inferred that the accuracy of these algorithms change significantly after a critical video quality.
Davis et al.~\cite{6556371} present a modular and adaptive algorithms for facial recognition on commercial off the shelf UAVs . They use human visual based approach with LBP features to train classifiers for facial recognition systems. However such systems are not immune to the large pose variations of faces as seen by the UAVs.

\vspace{-0.7cm}
\subsection{Commercially available drones that can track and follow}
\vspace{-0.4cm}
Currently, commercial following drones, such as Airdog~\cite{airdog}, Bebop \cite{bebop}, Hexo+~\cite{hexo} are popular. 
Airdog uses a wearable device called AirLeash to track and follow people. AirLeash can track movement and send control commands to AirDog.
Bebop has "Follow me" feature by using visual recognition technology and GPS tracking system on the smart phone. 
Same as Bebop drone, Hexo also uses smart phone to control and make it track and follow people.
For all these tracking drones, the users have to wear a tracking device to ensure that drones follow them, they are incapable autonomously capturing face images with good quality.
\vspace{-0.8cm}

\section{Proposed Approach}
\label{sec:approach}
\vspace{-0.4cm}
Face detection is the process of identifying image areas containing faces. Face recognition refers to matching the detected face to the reference. Thus face detection is the initial stage of recognition. The first limitation for a good face detection is the range of  detection. The size of the faces in images restrict the detections and faces seen from far are not detectable. To check the performance of face detection at different distances, we performed an empirical study similar to~\cite{hsu2015face} and obtained similar results.
                   
\begin{figure}
  \centering
  \includegraphics[width=0.7\textwidth]{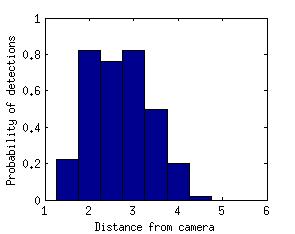}
  \caption{ Probability histogram for person detections at various distances from \SI{1}{\metre} to \SI{6}{\metre}}.
   \label{fig:prob}
\end{figure}

Specifically, in Figure~\ref{fig:prob} we plot the probability of obtaining a detection with respect to the distance maintaining a constant height of eye level. It can be observed that the performance degrades quickly after about \SI{3}{\metre}. In order to navigate the UAV into the zone of face detections we employ person following approach described in the preceding sections. 
\vspace{-0.6cm}
\vspace{-0.7cm}
The second limitation for obtaining a good face image is posed by the quality of the detected face. According to a study~\cite{nuero}, the recommendation for  face recognition in images are 32 pixels minimal distance between the eyes and 64 pixels for a better accuracy. These distances are determined by the quality of images as well as the distance at which they were captured. Although there have been efforts to increase the quality of sensors on UAVs, streaming with limited bandwidth introduces compression artifacts. The individual video frames and images are also affected by the velocity and movement of the sensor due to motion artifacts. Face detection algorithms perform reliably until a critical value of compression below which they degrade significantly~\cite{Korshunov:2011:VQF:2000486.2000488}. Further changes in illumination and pose variations make the process more challenging. In this paper we aim to minimize the pose variations seen by the UAV. Through continuous assessment of facial pose, we adjust the flight path to obtain an optimized view of face for a given image quality once it is above the face detection threshold.

\begin{figure}
  \centering
  \subfloat[Overview of the active face alignment system. The video feed and the odometry readings are the sensor readings obtained from Bebop. The input to the UAV are the control commands generated by a PD controller.]{ \includegraphics[width=0.7\textwidth]{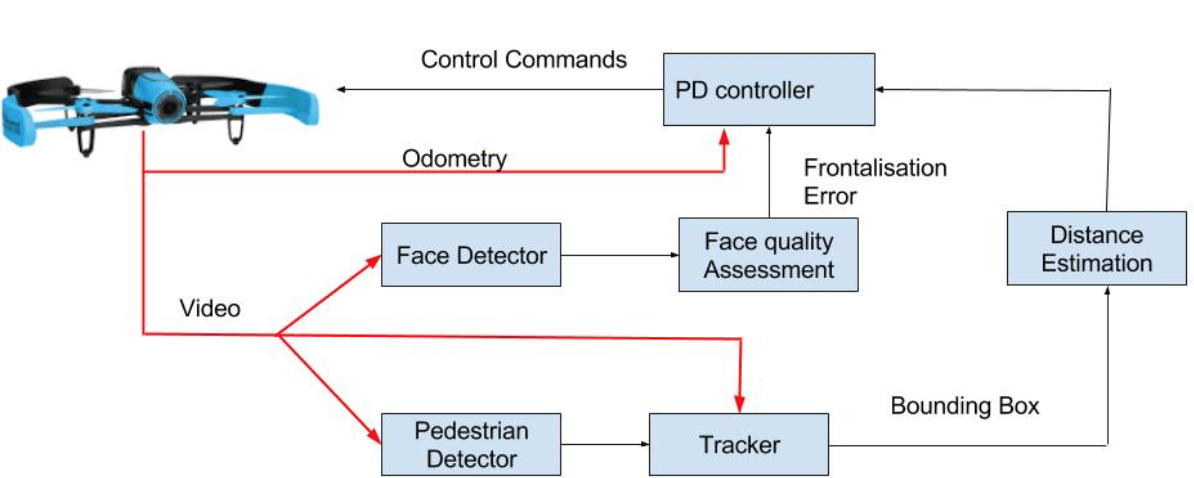}\label{fig:flowchart}}%
  
  \qquad
  \newline
  \subfloat[Table listing the frequencies of various nodes during the flight ]{%
    \begin{tabular}{| p{0.2\textwidth} p{0.2\textwidth}  p{0.3\textwidth}  p{0.2\textwidth}  p{0.2\textwidth} |} 
      \hline
      \\
      	Onboard Images & Odometry & Pedestrian Detector & Tracker & Face detector \\ \hline
        \\
        \SI{1.75}{\hertz} & \SI{5}{\hertz} & \SI{0.9}{\hertz} & \SI{1.4}{\hertz} &\SI{1.4}{\hertz}\\ \hline
      
    \end{tabular}
    \label{table}
  }
  \caption{System Overview}
  \label{fig:overview}
\end{figure}

The overview of our algorithm is described in Figure~\ref{fig:flowchart}. The communication from the UAV is established using ROS (Robot Operating System). The detector nodes subscribe to the high resolution onboard video feed. The frequencies of all the components of the architecture is listed in~\ref{table}. The pedestrian detector is concatenated with the tracker to increase the frequency of person detections. The bounding box obtained from the person detection is used to approximate the distance of the camera from the person. The PD controller issues the control signals based on the estimated distance to drive the UAV into the zone of face detections. For every detection obtained by the face detector node, an assessment on the face quality in terms of pose is performed and a frontalisation error is calculated. The error is used to navigate the UAV to th region of optimized face detections.
\vspace{-0.7cm}
\subsection{System Design}
\begin{enumerate}
\vspace{-0.4cm}
\item Parrot Bebop : Out of the several commercially available UAVs, Bebop Parrot is relatively safe to operate around humans. It has a  dual core processor with quad-core GPU, 8GB flash memory, and GPS. Connectivity is via Wi-Fi, and max operating distance is 1 mile. Therefore, it is mainly designed for outdoor flight. 
The on-board camera has 14 mega-pixels with a fish-eye lens. The on-board image resolution is $1920\times 1080$ (1080P). The streamed video has a resolution of $640\times 368$. Since the image quality is a critical factor, we resort to onboard images that have lower frequency but higher resolution. Although the high definition onboard images lowers the frame rate, the detection rate doesn't scale down. With greater resolution the chances of obtaining a detection on every frame increases. Bebop is a hobby grade UAV and  is not very bulky. It also comes with safety bumpers making it suitable for indoor flight.     
\item Data collection:   
We performed data collection in two stages. For the first stage, we collected data for drawing the correlation between the bounding box height obtained from person detection to the distance of person from the camera.  The ground truth for the distance was obtained using  April tag~\cite{olson2011tags} attached to the person. The UAV was manually moved around the person at different distances and the corresponding detections were recorded. 
In the second set of experiments, the same procedure was repeated but now the UAV was moved in concentric circles with varying radius. For different orientations of the camera from the frontal face and at different radii, the average frontalization error was obtained. Further, in order to compare our results with recognition accuracy, we used the probabilistic study on face recognition using Kinect 2.0 RGBD camera. The orientation of camera ranging from  $-90$$^{\circ}$ to $+90$$^{\circ}$ from the frontal face and radius up to \SI{5}{\metre} was covered. The performance accuracy of face recognition using Fischer faces \cite{Belhumeur:1997:EVF:261506.261512} was evaluated for the various poses. 

\end{enumerate}
\vspace{-0.7cm}

\vspace{-0.4 cm}

\subsection{Person Detection and Tracking } \label{hum det}
\begin{figure}
  \centering
  \includegraphics[width=0.7\textwidth,height=0.3\textheight]{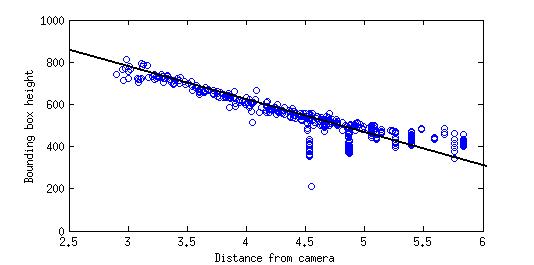}
  \caption{ Height of bounding box obtained from the pedestrian detection observed at various distances of person from the camera mounted on UAV. The correlation of the bounding box height and the distance is approximated to be linear ignoring the outliers. }
   \label{fig:depthheight}
\end{figure}

\vspace{-0.4cm}
Person detection and tracking using vision based cues has received some research attention in the past. With the focus on making identification more reliable, in this work we mainly design a simple approach using off-the-shelf, open-source detectors. The pre-trained pedestrian detector in OpenCV~\cite{pedestrian} is used detect people even in cluttered environment. The HoG features trained on linear SVM detects people in a video stream with very few false positives. However, the detector has relatively lower frequency compared to the incoming video stream as listed in {Table~\ref{table}}. Therefore, in order to bridge the gap between two detectors, we use object tracking. Many state-of-the-art trackers have been proposed previously~\cite{babenko2009visual,felzenszwalb2010object,kalal2012tracking,ma2015hierarchical}. However, for deformable objects, e.g. humans, it is still a challenging problem. In this work, we employ~\cite{nebehay2015clustering} for tracking people. CMT is a key-points based deformable part model. A set of keypoints are initially selected from the object to be tracked. At every time instant, the keypoints from the previous frames are matched and  the consensus of the points are used select object being tracked. CMT  is initialized with bounding box from the detector as shown in Figure~\ref{fig:flowchart} . After every detection, CMT is updated with the bounding box. This cascade of the detector and tracker works efficiently with very few false detections. In order to translate the position of person in image frame to the UAV frame, we make use of the dimensions of the bounding box. The center of the bounding box is aligned with the center of the image thus keeping the person in line of sight. Further, to estimate the depth of the camera from the person, we make use of the height of the bounding box. 
In Figure~\ref{fig:depthheight}, we plot the correlation of depth with the height of the detected bounding box using the ground truth from the AprilTags. The correspondences of the box height with the depth is approximated to be linear. At any instant of time, the estimation of depth is used to obtain the heading. Thus the UAV can be lead into the zone of face detections. 

\vspace{-0.7cm}





\vspace{-0.3cm}
\subsection{ Active Face Quality Assessment } 
\label{visibility}
\vspace{-0.4cm}
The general automated face recognition pipeline involves face detection, face alignment, feature extraction and feature matching~\cite{facerecognition}. In a realistic scenario, the chances of detected faces to be frontal are slim. Therefore, the pose variant faces are synthesized to give a frontal view of the face. Tal Hassner et.al.~\cite{DBLP:journals/corr/HassnerHPE14} propose a face frontalization technique that uses a fixed reference 3D model. The open source face detector from the DLIB library~\cite{dlib09} is used for cropping faces from the background. As we focus on images captured from relatively low altitudes, we can safely claim that the HoG-based features can distinguish human faces through contrast. The frontalization is carried out by first detecting certain facial features and projecting the corresponding pixel intensities  from 2D space to fixed reference 3D model of face. In order to account for the  occlusions and pose variations, the authors use visibility scores originally used by 3D reconstruction methods~\cite{791235},~\cite{ZPQS05}. The visibility score for each pixel in the projected 3D surface $q$ is then estimated to be:
\begin{equation}
v(q)=1-exp(-N_q)
\label{visibility}
\end{equation}
Here $N_q$ is the number of times a query pixel $q$ is accessed while forming correspondence of the 2D query image pixel with the reference 3D surface.  As the face turns away from the camera increasing the angle, fewer pixels in the query image are mapped to the 3D pixels, hence reducing the visibility of the region facing away.
\begin{figure}
  \centering
  \includegraphics[width=0.8\textwidth]{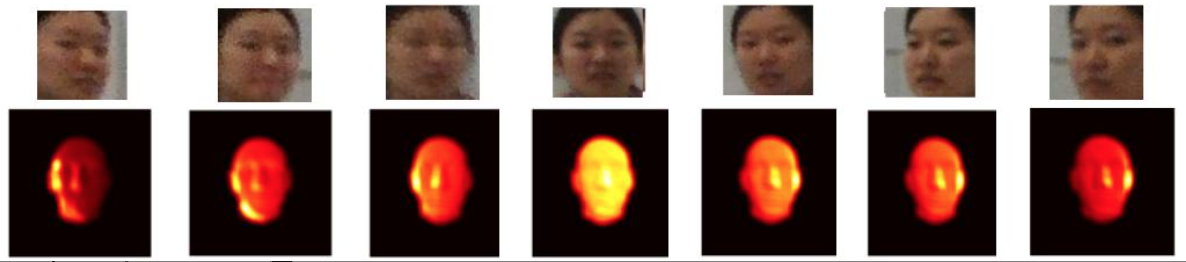}
  \vspace{-0.2cm}
  \caption{Heat map showing the visibility at each pixel in different orientations of face.}
   \label{fig:heatmap}
\end{figure}     

Figure~\ref{fig:heatmap}. displays the visibility on surface of faces in three different poses captured during flight. The face detection algorithms renders a tight bounding box around all the pixels that constitute face. Since the frontalization is preceded by the person following controller, we make sure that the center of the detected face is aligned to the center of the UAV frame though visual servoing.  The gradient of visibility gives a crude estimate regarding the part of the face facing the camera. We define a frontalization error as the difference between the average visibility scores of right half pixels and average visibility scores of left half pixels of face. Thus the frontalisation error is :
\begin{equation}
    Frontalisation~error = \frac{\sum_{q=R_q}^{} v_q}{R_q} -\frac{\sum_{q=L_q}^{} v_q}{Lq}
\end{equation}

Here $R_q$ are the right pixels and $L_q$ are the left pixels.

\vspace{-0.7cm}
\subsection{Face Verification}
\vspace{-0.4cm}
In order to analyze the effectiveness of the proposed approach in the face recognition pipeline, we perform face verification along the flight path. The faces captured throughout the flight path must show an increasing similarity to the frontal face. Facial biometrics are highly expressive and diverse. Convolutional neural networks show capacity to deliver an intricate representation of images. The deep cascade of layers give a high level representation of faces. This representation of faces can be used for a one-to-one comparison of face against a registered face. In this paper, we used the Deep CNN architecture VGG-16 proposed by~\cite{DBLP:journals/corr/HassnerHPE14} to extract the features from the detected face.

\begin{figure}
  \centering
  \includegraphics[width=0.8\textwidth]{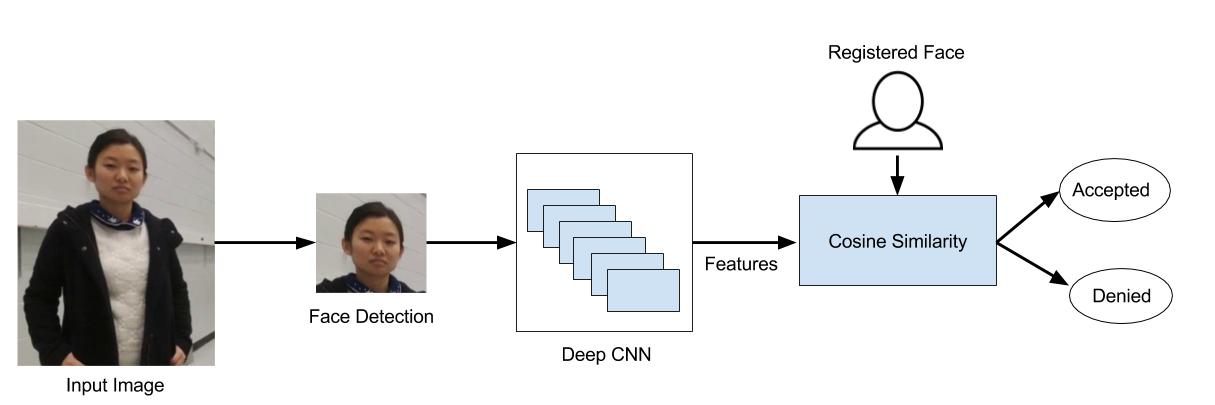}
  \caption{A generic CNN based verification system consists of face detection,face alignment, feature extraction followed by similarity matching}
   \label{fig:face_ver}
\end{figure}

\begin{figure}
  \centering
  \includegraphics[width=0.7\textwidth]{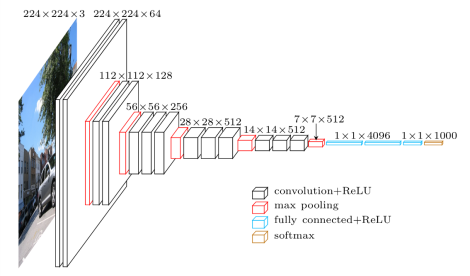}
  \caption{The figure describes the VGG-16 architecture, a network used to extract features for faces from a high dimensional representation of images.}
   \label{fig:vgg}
\end{figure}

The architecture as depicted in Figure~\ref{fig:vgg} consists of series of convolutional layers and max-pooling layers followed by the fully connected layer and soft-max layer. The convolutional layer extracts features using a sliding window convolving over patches of the image. The intermediate pooling layers are used to sub sample the features. The last three fully connected layers are similar to the general neural network architecture. The network is pre-trained on ImageNet dataset~\cite{imagenet_cvpr09} consisting of 15 million images of various objects. The features from the third fully connected layer is used for representing face. In biometrics, a similarity measure between the registered image and the query image is used for verification. Higher similarity of the query face indicates more confidence of the person being verified. The cosine similarity  for any two vectors A and B  is calculated as follows:
\begin{equation}
CS(A,B)=\frac{A.B}{||A|| ||B||}
\label{cosine similarity}
\end{equation}
The face verification scheme based on cosine similarity is represented in Figure~\ref{fig:face_ver}. Face recognition is simply  face verification over all the registered users. However in our experiments, we perform face verification for a single user as the multi-user verification requires a large database of registered users.
\vspace{-0.7 cm}
\vspace{-1cm}

\vspace{-1cm}
\subsection{Vision Based Controllers} 
\label{controller}
\begin{figure}
  \centering
  \subfloat[The architecture for the proposed person following controller ]{ \includegraphics[width=0.8\textwidth]{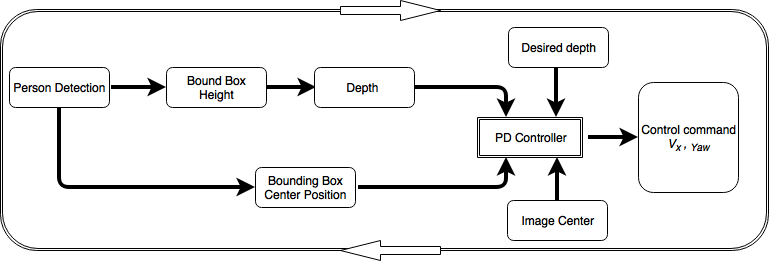}\label{fig:controller1}}%
  \qquad
  \newline
\subfloat[The controller architecture for the frontalisation error based visual servoing. The cascade of controllers is represented by a high frequency  inner odometry loop and a low frequency outer observation loop.]{ \includegraphics[width=0.8\textwidth]{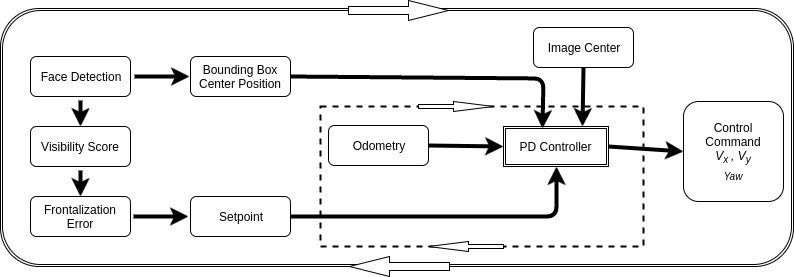}\label{fig:controller2}}
  \caption{The overview of the vision based controllers}
  \label{fig:controller}
\end{figure}

\vspace{-0.4cm}
An overview of the system is shown in Figure\ref{fig:controller}. The control architecture consists of two distinct parts that both take visual cues from the on-board camera to produce motion commands. When no faces are detected, the first controller looks for people in the image and approaches them. Once the UAS is close enough to reliably detect a face, a fontalization controller takes over.

\vspace{-0.7cm}
\subsubsection{Person Following Controller}
\vspace{-0.3cm}
Our person following module navigates the UAV into the region with higher probability of face detection. The first person detection seen by the UAV initiates the take-off. After the initialization, the aim is to head towards the person keeping the person in the center of the frame. The upper segment of the controller illustrates the person following controller. From the bounding box detected, two features are calculated, the height of the bounding box and the center of the bounding  box. The height is used for estimating the depth as explained in Section~\ref{hum det}. The UAV moves progressively towards the person with a continuous estimation of depth. As UAV flies directly at the person, we always maintain a minimum distance of \SI{1.5}{\metre} for safety. Further, to center the UAV with respect to the person, we  align the center of the detection with the center of the image in the pixel space by issuing the yaw commands. Person following thus reduces the distance of the camera from the person increasing the chances of face detections. In the next section, we define the controller design for active face frontalization.
 
\vspace{-0.7cm}
\subsubsection{Visual Servoing on Frontalization Error}
\vspace{-0.3cm}
Owing to the small range of operation for the active face frontalization, the inputs for the face frontalization unit are a set of waypoints through the course of flight path.
The system with a cascade of controllers is illustrated in the Figure~\ref{fig:controller2}.  
From the detected face, the frontalization error is calculated as explained in Section~\ref{visibility}. A setpoint or the target point for the UAV is set based on the fronatlization error. Based on the current odometry of the UAV and the desired setpoint, the PD controller issues velocities about the longitudinal and lateral axis of the UAV. However, the frequency of the odometry is much higher than the face detections. Therefore, in the inner loop, with every estimate of current odometry and the desired setpoint, the PD controller issues a new velocity command. However the setpoint is updated as and when a face is detected in the outer loop.  The outer loop also sets the yaw for the UAV creating a circling movement around the face. The yaw is issued based on the difference between the center of the face and the center of the image. This way the UAV is oriented towards the detected face.
\vspace{-0.5cm}

\section{Results}
\begin{figure}[!h]
\begin{minipage}[b]{.4\linewidth}
 \centering
 \centerline{\includegraphics[width=4cm, height=3cm]{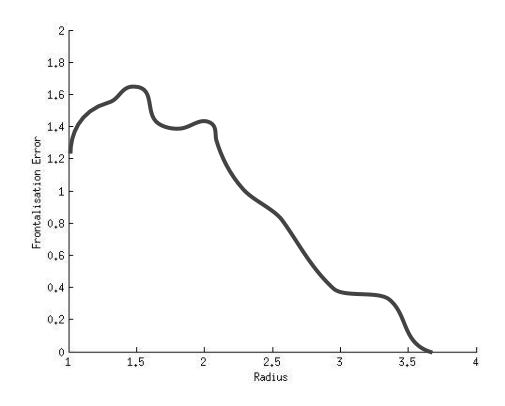}}
  \vspace{-0.2cm}
 \centerline{(a)}\medskip
\end{minipage}
\hspace{0.5cm}
\begin{minipage}[b]{0.4\linewidth}
 \centering
 \centerline{\includegraphics[width=4cm, height=3cm]{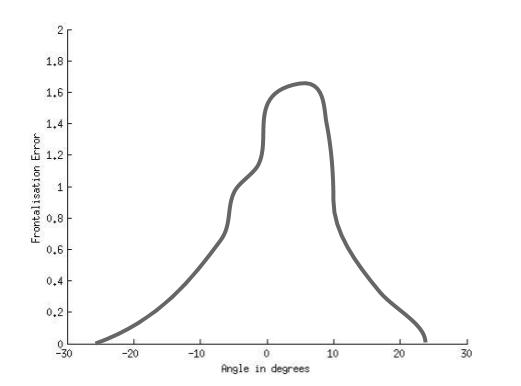}}
 \vspace{-0.2cm}
 \centerline{(b)}\medskip
\end{minipage}
\vfill
\begin{minipage}[b]{0.4\linewidth}
 \centering
 \centerline{\includegraphics[width=4cm, height=3cm]{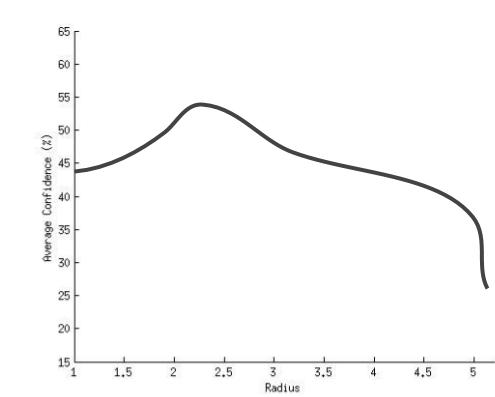}}
\vspace{-0.2cm}
 \centerline{(c)}\medskip
\end{minipage}
\hspace{0.5cm}
\begin{minipage}[b]{0.4\linewidth}
 \centering
 \centerline{\includegraphics[width=4cm, height=3cm]{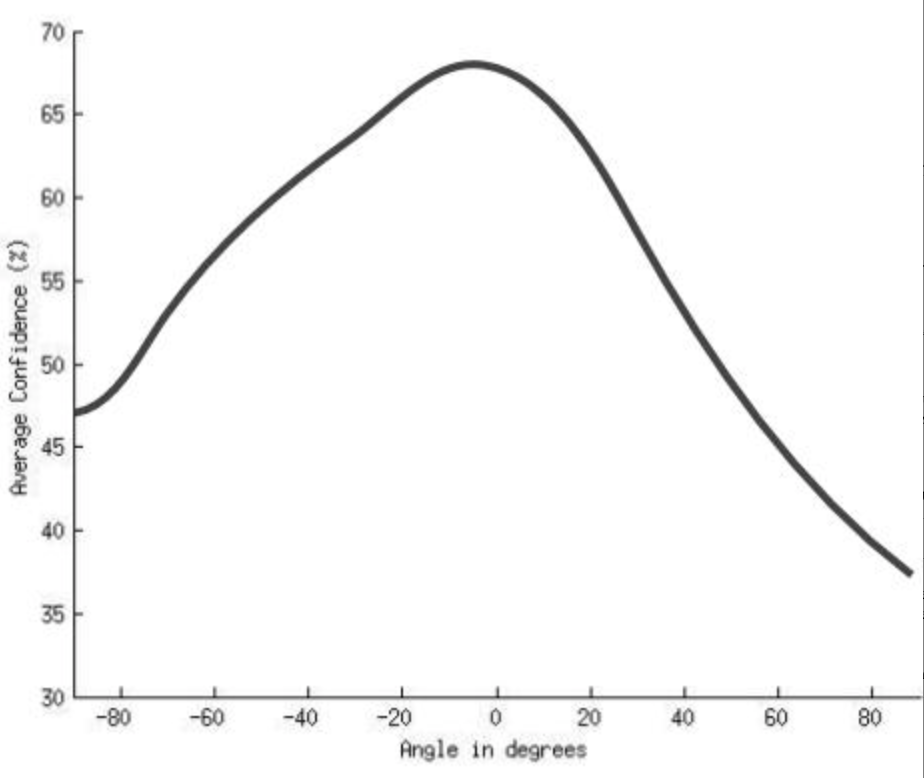}}
 \vspace{-0.2cm}
\centerline{(d)}\medskip
\end{minipage}
\caption{(a) Frontalization accuracy vs radius for on board camera, (b) Frontalization accuracy vs orientation for onboard camera, (c) Average Recognition confidence vs radius from Kinect and (d) Average Recognition confidence vs orientation from Kinect}
\label{fig:visib_plot}
\end{figure}

\vspace{-0.3cm}
\label{sec:eval}
In order to asses frontalization error as a quantitative measure of the face detection, we perform several experiments. Firstly, we fly the UAV at different angles and distances from the person. For each of the sampled points, frontalization accuracy is calculated as:
$$Frontalization~accuracy = 1-Frontalisation~error$$
In Figure~\ref{fig:visib_plot} (a) and (b) frontalization accuracy as a function of distance and orientation from the face is plot. Since the frontalisation process begins after aligning the facial bounding box to the center of the UAV frame, experiments of frontalisation error with respect to the altitude of UAV is not relevant here. During the process of frontalisation, the altitude of UAV with respect to face bounding box is maintained constant.

 A visual representation of the velocity vectors is shown Figure \label{fig:vel}. At every location on the field of experiment we represent the strength and direction of the velocity using arrows. all the vectors point towards the human positioned at [0,0.5]. The velocities are calculated with respect to the relative position of UAV with the person detection bounding box as described in Section \ref{hum det}
We compare our results with the empirical data obtained by performing a similar analysis on Kinect camera for face recognition. Figure~\ref{fig:visib_plot} (c) and (d) is a plot of recognition accuracy with respect to orientation and distance observed from the Kinect camera. Figure~\ref{fig:visib_plot} establishes a close correspondence of frontalization error with the face recognition confidence.

The accuracy is high when the camera is oriented at $0$ $^{\circ}$ from the face and subsides down at higher angles leading to a more severe pose of face. A similar result was observed for the radius. 
As explained in Section \ref{controller}, we use a PD controller to navigate the UAV to the center. 

\begin{figure}[h!]
  \centering
  \includegraphics[width=0.5\textwidth]{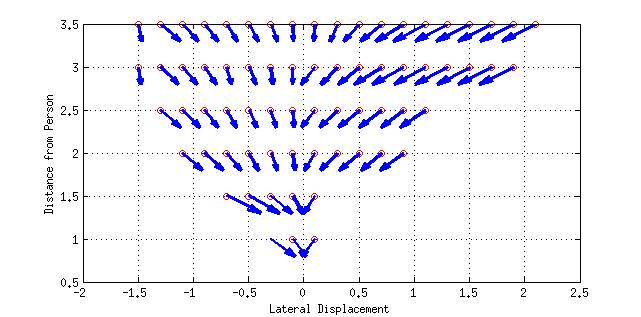}
 \captionsetup{justification=centering}
  \caption{The plot visualizes the velocity commands that are calculated at each location with respect to the detected faces.}
   \label{fig:vel}
\end{figure}    

In Figure~\ref{fig:vel}, the control signals generated using the frontalization error at any given point in the space is shown. The arrows converge towards the \textit{dead zone}, where the frontalization error is zero or the view of face is fully frontal.

\begin{figure}[h!]
  \centering
  \includegraphics[width=0.4\textwidth]{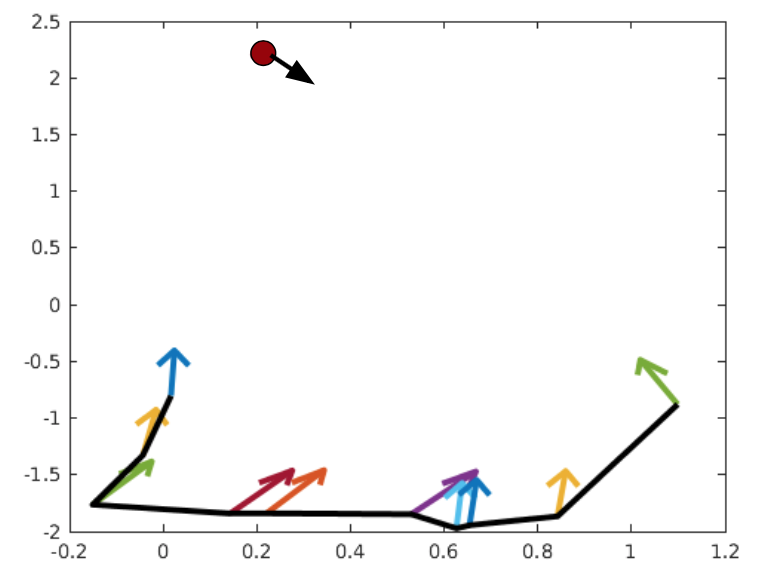}
  \captionsetup{justification=centering}
  \caption{UAV trajectory for a single run. The color map indicate the time lapse. The person is standing at the origin looking straight }
   \label{path}
\end{figure}   


\begin{figure}
  \centering
  \subfloat[Frontalization error during the flight path. The error saturates at zero when a frontal view is obtained ]{ \includegraphics[width=0.4\textwidth]{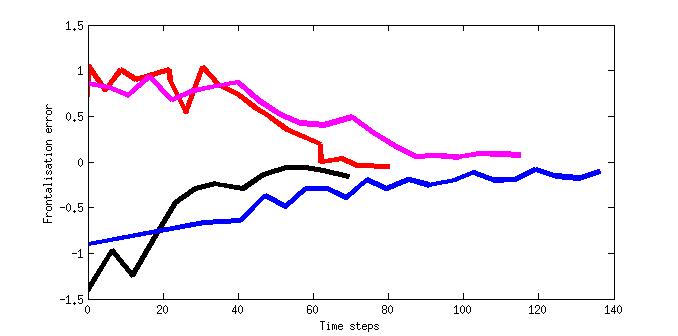}\label{fig:vis_traj}}%
  \qquad
\subfloat[Similarity measure for the face detected along the proposed flight path.]{ \includegraphics[width=0.4\textwidth]{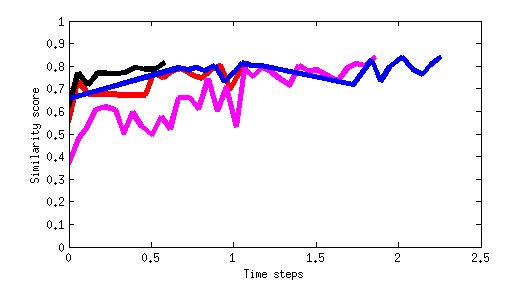}\label{fig:face_score}}
  \caption{The trajectories of frontalisation error and cosine similarity as a function of time steps}
  \label{fig:trajectories}
\end{figure}
 \vspace{-0.5cm}

 

We test the interface described in Section~\ref{controller} in an indoor setting.  A typical path traversed during active face frontalization is shown in the Figure~\ref{path}. The position of the person and the orientation is marked by the red circle and the arrow respectively. The yaw of the UAV at every instant is represented along the path. By taking off at different angles away from the face, we track the frontalization error along the trajectory. As shown in Figure~\ref{fig:vis_traj}, the error starts off at a high value, with positive error indicating the right side of face and negative error indicating the left part. The error gradually converges to zero. Any drift is efficiently sustained. 
In the final stage of evaluation, to justify our claim that active frontalization boosts the performance of face recognition, cosine similarity scores for the faces detected along  flight path and the registered face for different runs is  calculated. In Figure~\ref{fig:face_score}, it can be observed that the similarity scores for the images along the trajectory increases along the flight path. The higher similarity yields better face recognition.
\vspace{-0.7cm}

 

\section{Conclusion}
\vspace{-0.2cm}
The contribution of this paper is twofold. Firstly, we provide an autonomous biometric information gathering system based on UAVs. Although literature suggest research aiming to incorporate face technology with UAVs. We make the first attempt of actively assessing biometric data on UAVs as it is gather and to improve the quality of the gathered information. We make use of soft biometric cues to implement autonomy of motion useful to a particular class of applications  like UAV surveillance and photography.
Secondly, we introduce frontalization error derived from the existing methods that can be used online for adjusting flight paths to improve the image quality to aid verification. Through experimentation we show that the proposed approach leads to a better quality of faces images for recognition. 
We present a inexpensive consumer system that is modular and can be adapted to any ROS enabled UAV platform. The software components are also off-the-shelf and could likely be adapted improve the overall system performance. 
\vspace{-0.3cm}
\begin{acknowledgement}
This material is based upon work supported by the National Science Foundation under Grant IIP \#1266183. We are grateful to the CSE Department of University at Buffalo for supporting Nagashri N L and Yifang Liu.
\end{acknowledgement}
\vspace{-0.8cm}

\bibliographystyle{acm}

\bibliography{ref}

\backmatter
\appendix
%
%
%

\chapter{Chapter Heading}
\label{introA} 

Use the template \emph{appendix.tex} together with the Springer document class SVMono (monograph-type books) or SVMult (edited books) to style appendix of your book in the Springer layout.

\section{Section Heading}
\label{sec:A1}
Instead of simply listing headings of different levels we recommend to let every heading be followed by at least a short passage of text. Further on please use the \LaTeX\ automatism for all your cross-references and citations.

\subsection{Subsection Heading}
\label{sec:A2}
Instead of simply listing headings of different levels we recommend to let every heading be followed by at least a short passage of text. Further on please use the \LaTeX\ automatism for all your cross-references and citations as has already been described in Sect.~\ref{sec:A1}.

For multiline equations we recommend to use the \verb|eqnarray| environment.
\begin{eqnarray}
\vec{a}\times\vec{b}=\vec{c} \nonumber\\
\vec{a}\times\vec{b}=\vec{c}
\label{eq:A01}
\end{eqnarray}

\subsubsection{Subsubsection Heading}
Instead of simply listing headings of different levels we recommend to let every heading be followed by at least a short passage of text. Further on please use the \LaTeX\ automatism for all your cross-references and citations as has already been described in Sect.~\ref{sec:A2}.

Please note that the first line of text that follows a heading is not indented, whereas the first lines of all subsequent paragraphs are.

%
\begin{figure}[t]
\sidecaption[t]
\includegraphics[scale=.65]{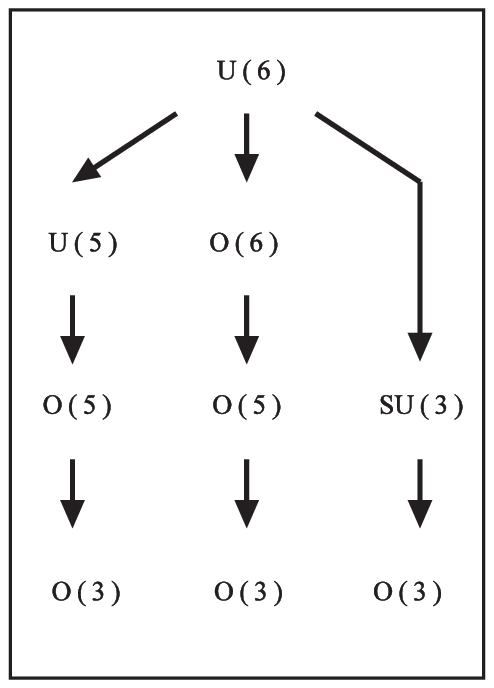}
%
%
\caption{Please write your figure caption here}
\label{fig:A1}       
\end{figure}

%
\begin{table}
\caption{Please write your table caption here}
\label{tab:A1}       
%
%
\begin{tabular}{p{2cm}p{2.4cm}p{2cm}p{4.9cm}}
\hline\noalign{\smallskip}
Classes & Subclass & Length & Action Mechanism  \\
\noalign{\smallskip}\hline\noalign{\smallskip}
Translation & mRNA$^a$  & 22 (19--25) & Translation repression, mRNA cleavage\\
Translation & mRNA cleavage & 21 & mRNA cleavage\\
Translation & mRNA  & 21--22 & mRNA cleavage\\
Translation & mRNA  & 24--26 & Histone and DNA Modification\\
\noalign{\smallskip}\hline\noalign{\smallskip}
\end{tabular}
$^a$ Table foot note (with superscript)
\end{table}
%

%
%

\Extrachap{Glossary}

Use the template \emph{glossary.tex} together with the Springer document class SVMono (monograph-type books) or SVMult (edited books) to style your glossary\index{glossary} in the Springer layout.

\runinhead{glossary term} Write here the description of the glossary term. Write here the description of the glossary term. Write here the description of the glossary term.

\runinhead{glossary term} Write here the description of the glossary term. Write here the description of the glossary term. Write here the description of the glossary term.

\runinhead{glossary term} Write here the description of the glossary term. Write here the description of the glossary term. Write here the description of the glossary term.

\runinhead{glossary term} Write here the description of the glossary term. Write here the description of the glossary term. Write here the description of the glossary term.

\runinhead{glossary term} Write here the description of the glossary term. Write here the description of the glossary term. Write here the description of the glossary term.
\printindex


\end{document}